\documentclass[10pt,twocolumn,letterpaper]{article}
\usepackage[pagenumbers]{cvpr} 
\def\paperID{12295} 
\def\confName{CVPR}
\def\confYear{2025}

\title{Training-free Dense-Aligned Diffusion Guidance for Modular Conditional Image Synthesis}

\author{
Zixuan Wang$^{1}$ \quad
Duo Peng$^{2}$ \quad
Feng Chen$^{3}$ \quad
Yuwei Yang$^{4}$ \quad
Yinjie Lei$^{1, *} $
\\
$^1$Sichuan University \quad
$^2$Singapore University of Technology and Design \\
$^3$The University of Adelaide \quad
$^4$Australian National University
\\
{\tt\small zixuan98@stu.scu.edu.cn, duo\_peng@mymail.sutd.edu.sg, chenfeng1271@gmail.com} \\
{\tt\small yuwei.yang@anu.edu.au, yinjie@scu.edu.cn}
}

\begin{document}
\maketitle
\renewcommand{\thefootnote}{} 
\footnotetext{
* Corresponding Author: Yinjie Lei (yinjie@scu.edu.cn)
}

\begin{abstract}
Conditional image synthesis is a crucial task with broad applications, such as artistic creation and virtual reality. However, current generative methods are often task-oriented with a narrow scope, handling a restricted condition with constrained applicability. In this paper, we propose a novel approach that treats conditional image synthesis as the modular combination of diverse fundamental condition units. Specifically, we divide conditions into three primary units: text, layout, and drag. To enable effective control over these conditions, we design a dedicated alignment module for each. For the text condition, we introduce a Dense Concept Alignment (DCA) module, which achieves dense visual-text alignment by drawing on diverse textual concepts. For the layout condition, we propose a Dense Geometry Alignment (DGA) module to enforce comprehensive geometric constraints that preserve the spatial configuration. For the drag condition, we introduce a Dense Motion Alignment (DMA) module to apply multi-level motion regularization, ensuring that each pixel follows its desired trajectory without visual artifacts. By flexibly inserting and combining these alignment modules, our framework enhances the model's adaptability to diverse conditional generation tasks and greatly expands its application range. Extensive experiments demonstrate the superior performance of our framework across a variety of conditions, including textual description, segmentation mask (bounding box), drag manipulation, and their combinations. Code is available at \textcolor[RGB]{237,0,140}{https://github.com/ZixuanWang0525/DADG}
\end{abstract}
    
\section{Introduction}
\begin{figure}[tbp]
    \centering
    \begin{subfigure}{0.47\textwidth}
        \includegraphics[width=\linewidth]{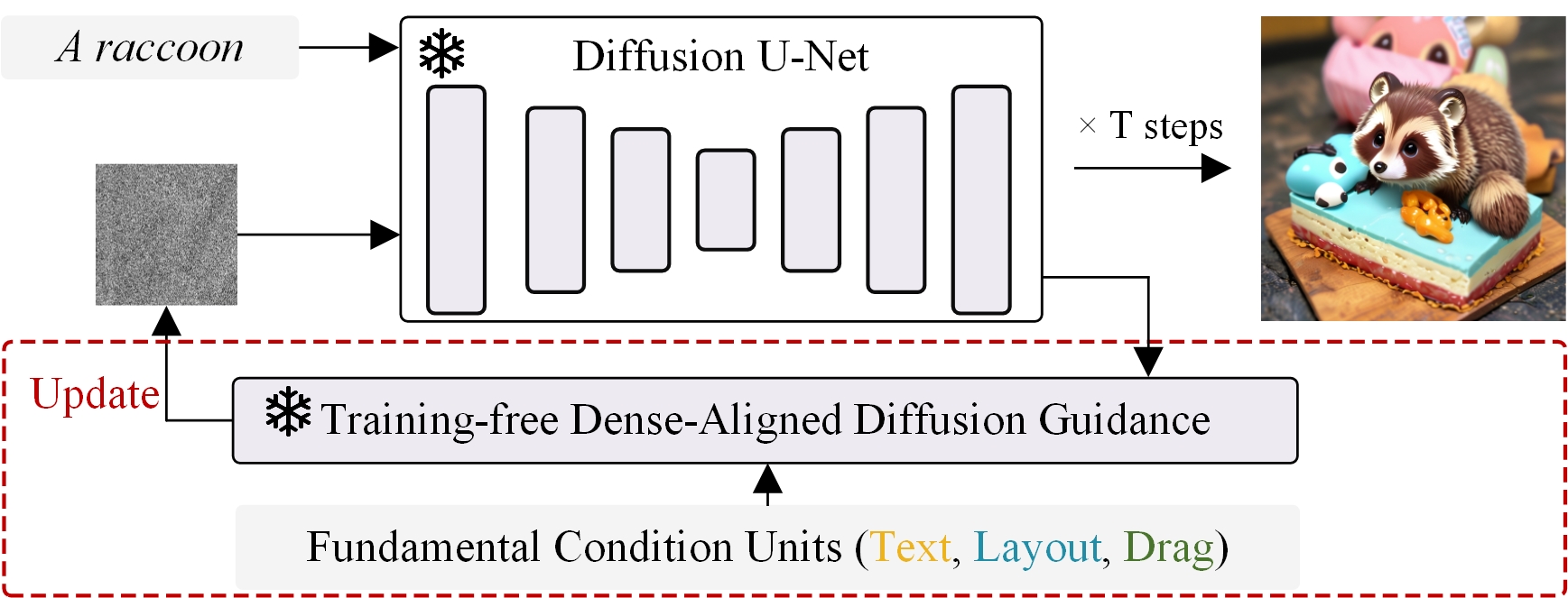}
        \caption{General pipeline of our proposed modular conditional image synthesis. Overall framework of Training-free Dense-Aligned Diffusion Guidance is illustrated in Fig. \ref{fig:sub2}. }
        \label{fig:sub1}
    \end{subfigure}
    \hfill
    \begin{subfigure}{0.47\textwidth}
        \includegraphics[width=\linewidth]{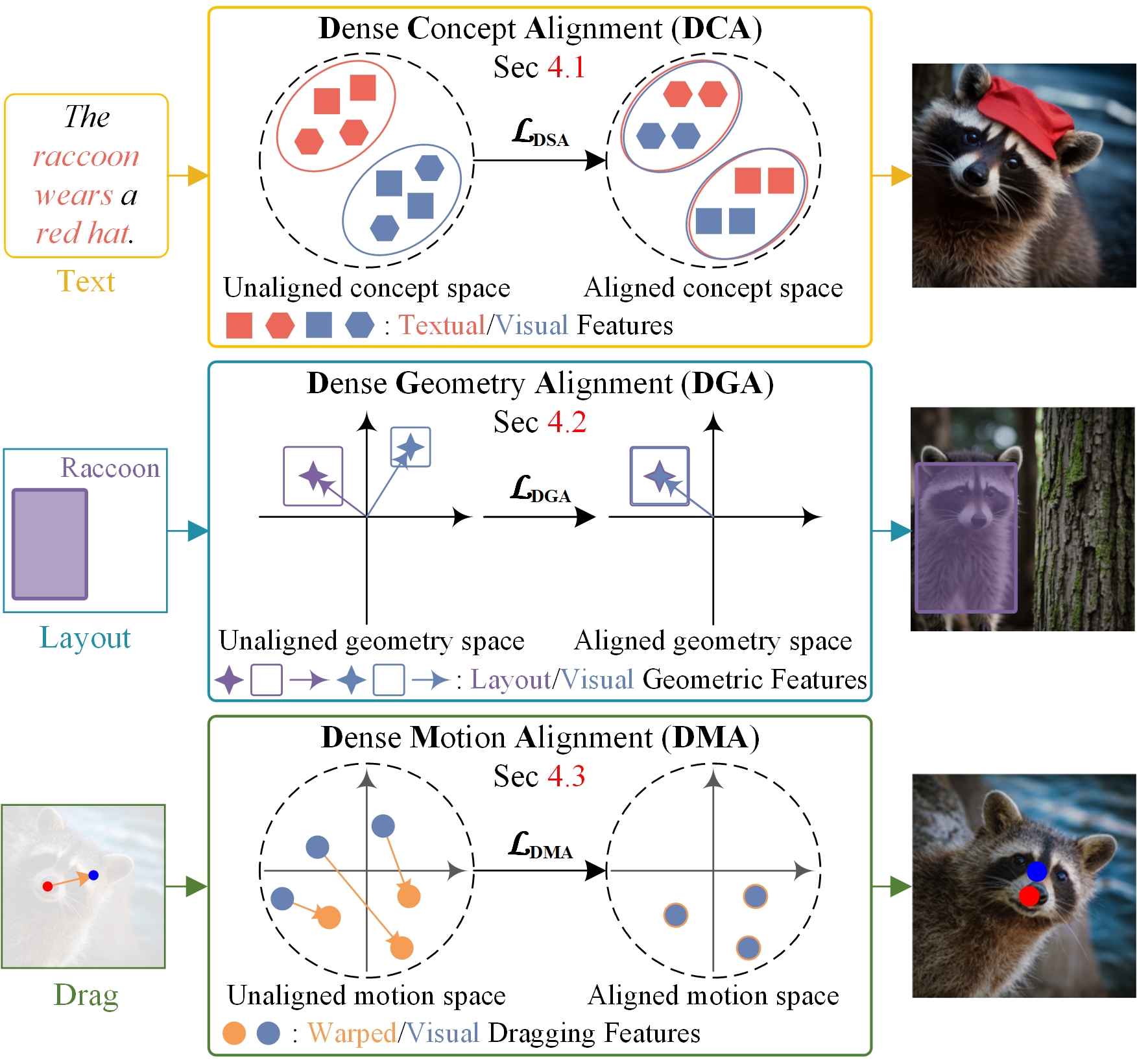}
        \caption{Illustration of our proposed Training-free Dense-Aligned Diffusion Guidance. We design three dense alignment modules to enable an effective accommodation of each specific condition unit.}
        \label{fig:sub2}
    \end{subfigure}
    \vspace{-10pt}
    \caption{The overview of our modular conditional image synthesis. Driven by modular combination of fundamental condition units, diverse visual content can be synthesized.}
    \vspace{-20pt}
    \label{fig:main}
\end{figure}
Conditional image synthesis aims to generate realistic images based on user-provided requirements, playing a pivotal role in various applications, such as artistic creation \cite{zeng2024jedi,ruiz2024hyperdreambooth} and virtual reality \cite{kim2023neuralfield, lei2023recent}. Recent advances in diffusion models have demonstrated promising performance in handling diverse conditional signals, including image descriptions \cite{tumanyan2023plug,rombach2022high}, segmentation maps \cite{kim2023dense,mou2024t2i,mo2024freecontrol}, bounding boxes \cite{chefer2023attend,xie2023boxdiff,zhao2024uni}, and drag information \cite{shi2024dragdiffusion,hou2024easydrag}, among others. While existing approaches are effective in interpreting a particular form of input from users, they struggle to generalize across a wide range of conditions. This is impractical for flexible generation of complex visual scenes in the real world.

In this paper, we propose the \textit{\textbf{M}odular \textbf{C}onditional \textbf{I}mage \textbf{S}ynthesis (\textbf{MCIS})} paradigm, which regulates the synthesis process by selectively applying and combining independent fundamental condition units, as demonstrated in Fig.~\ref{fig:main} (a). Specifically, we categorize condition units into: (1) \textit{Text.} This category refers to natural language descriptions, specifying the semantic content of an image; (2) \textit{Layout.} This category mainly encompasses segmentation maps and bounding boxes, representing the component arrangement and scene structure; and (3) \textit{Drag.} This category refers to point pairs (original and destination of drags), encoding the local transformations of an image. However, MCIS presents a non-trivial problem, as it requires ensuring that all conditions are distinctly reflected in the visual content without omission.

We sequentially analyze the key challenges involved in aligning each basic condition unit: (1) \textit{Concept Mismatching}. Text typically conveys detailed information about the properties of individual objects and their interactions. This requires generative models to match fine-grained region-level and word-level concepts, instead of considering the image and its description as a whole for alignment; (2) \textit{Geometry Inconsistency}. Layout implicitly encodes complex relationships among multiple objects, such as size ratios and relative positions. Ensuring image realism requires establishing a detailed geometric correspondence between synthesized content and spatial configuration; and (3) \textit{Motion Disharmony}. Drag often provides displacement vectors indicating the visual regions to be moved. To synthesize visually coherent content, it is crucial to effectively maintain the intended movement of each visual element, while preserving both appearance and semantic consistency. 

To address the above issues, we propose a dense-aligned diffusion guidance framework for modular conditional image synthesis. As illustrated in Fig.~\ref{fig:main} (b), our framework utilizes plug-and-play guidance modules to allow visual content to independently align with each of the condition units. Specifically, we propose the Dense Concept Alignment (DCA) module, which ensures the consistency between visual content and scene description in a coarse-to-fine manner. Beyond scene-level alignment, we establish fine-grained correspondence from attribute and relational perspectives within a disentangled feature space. We further introduce the Dense Geometry Alignment (DGA) module. Building upon the detection information from the synthesized image, it imposes constraints on object-wise locations, as well as on relative sizes and distances between each pair of objects. This process ensures the separation between different foreground instances while maintaining the realism of scene arrangement. Based on dense displacement fields, we design the Dense Motion Alignment (DMA) module. During diffusion sampling, this module guides the variation of visual content between adjacent timesteps by aligning it with the displacement fields. Meanwhile, it uses pixel-wise color and semantic regularization to achieve photorealistic consistency.

To evaluate the effectiveness of our framework in MCIS, we conduct extensive experiments on public benchmarks, including COCO \cite{lin2014microsoft}, DenseDiffusion \cite{kim2023dense}, DrawBench \cite{saharia2022photorealistic}, and DragBench \cite{shi2024dragdiffusion}. Quantitative and qualitative results demonstrate that our approach significantly improves the adherence to the text, layout, and drag conditions. Moreover, our guidance modules are compatible to simultaneously control over multiple condition units. In addition, our framework can be seamlessly integrated into various diffusion architectures. 

Our contributions are summarized as follows:
\begin{itemize}[left=1em, align=parleft]
\item We propose a paradigm of modular conditional image synthesis, which expands the application scope of diffusion models by the adaptive combination of fundamental condition units.
\item We design three plug-and-play dense alignment modules, DCA, DGA, and DMA, to independently achieve effective control over concept, geometry, and motion.
\item Extensive experiments show that our framework significantly enhances controllability across text, layout, drag, and their combinations.
\end{itemize}

\section{Related Works}
Conditional image synthesis is a core research field in AIGC, which aims to generate images explicitly guided by input conditions. From the aspect of framework architecture, current studies can be divided into: (1) \textit{Training-based} and (2) \textit{Training-free} condition image synthesis. Crucially, we classify diverse conditions into three fundamental units: text, layout, and drag. Training-based and -free frameworks are, thus, sequentially analyzed in a context of fundamental condition units we define.

\subsection{Training-based Conditional Image Synthesis}
Training-based frameworks often train diffusion models from scratch or lightweight auxiliary modules to accommodate user-provided conditions. \textit{Text-driven} frameworks, e.g., Stable Diffusion family \cite{rombach2022high,podellsdxl,esser2024scaling} and Pix-Art family \cite{chenpixart,chen2024pixart}, optimize diffusion models from scratch to implicitly aware of embeddings from frozen large-language models, such as CLIP \cite{radford2021learning} and T5 \cite{raffel2020exploring}. \textit{Layout-driven} approaches incorporate parameter-efficient modules to encode segmentation masks, bounding boxes, and depth maps, among others. For instance, GLIGEN \cite{li2023gligen} uses a learnable gated self-attention module to capture grounding information, \textit{i.e.}, regional bounding boxes and its corresponding phrase. ControlNet family \cite{zhang2023adding,zhao2024uni,li2025controlnet} can be viewed as a learnable copy of a U-net encoder, which reuses an off-the-shelf diffusion model as a backbone for handling diverse spatial conditioning vectors. The copy encoder is connected with U-net decoder layers in a frozen diffusion model by zero-convolution modules. Instead of a Siamese network of encoder layers, T2I-Adapter \cite{mou2024t2i} learns various low-complexity adapters according to different conditions, where each adapter is only composed of some residual convolution blocks. 
However, the demands for high-quality data and considerable computational resources remain the bottlenecks for scalability and widespread adoption of the above approaches.

\subsection{Training-free  Conditional Image Synthesis}
Training-free works focus on modifying the original diffusion sampling process to drive the conditional image synthesis using feedback from the guidance function. For \textit{text} condition, mainstream guidance frameworks are classifier and CLIP guidance. Specifically, classifier guidance \cite{dhariwal2021diffusion} incorporates class information into diffusion models using gradients of a timestep-aware image classifier. CLIP guidance, such as DOODL \cite{wallace2023end}, replaces classification scores with global CLIP scores, which can make generation process accommodate free-form descriptions. Regarding \textit{layout} condition, current approaches often implicitly modify distributions of attention scores to achieve guidance. The underlying assumption is: spatial distribution of high-response attention aligns perceptually with the locations of objects in synthesized images \cite{chefer2023attend}. Based on this idea, BoxDiff \cite{xie2023boxdiff} and R\&B \cite{xiao2023r} impose region and boundary constraints to stored cross-attention maps. DenseDiffusion \cite{kim2023dense} and A$\&$R \cite{phung2024grounded} further leverage semantic affinity in self-attention maps to ensure more accurate object placement. With respect to \textit{drag} condition, DragDiffusion \cite{shi2024dragdiffusion} and DragNoise \cite{liu2024drag} employ motion supervision and point tracking steps to optimize diffusion semantic features. These steps enable features of original points to move to destination points as effectively as possible. However, aforementioned approaches are often designed to encode a particular condition form, which restricts their generalization across a wide range of conditions.  
\section{Preliminary}
\label{sec:preliminary}
This section provides a brief overview of the core mechanisms in the conditional diffusion model framework, including diffusion sampling and diffusion guidance.

\textbf{Diffusion sampling.} Diffusion sampling is designed to generate a clean image by progressively removing noise from a random Gaussian noise image $ \mathbf{x}_{T}$ \cite{ho2020denoising,yu2023freedom}. This sampling process is accomplished by a learned denoising neural network $\epsilon_{\theta}$ which is conditioned on a given prompt $ \mathbf{y}$. Concretely, at each time step $t$, a cleaner image $ \mathbf{x}_{t-1}$ can be generated by subtracting the predicted noise $\hat{\epsilon}_{t}=\epsilon_{\theta}(\mathbf{x}_{t}; t, \mathbf{y}) $ from the current noise image $ \mathbf{x}_{t}$. As an example of denoising diffusion implicit model (DDIM) \cite{2021Denoising}, $ \mathbf{x}_{t-1}$ can be produced via:

\begin{equation}
\mathbf{x}_{t-1}=\sqrt{\alpha_{t-1}}\hat{\mathbf{x}}_{0}+\sqrt{1 - \alpha_{t-1}}\hat{\epsilon}_{t} \text{,}
\end{equation}
where
\begin{equation}
\hat{\mathbf{x}}_{0}=\frac{\mathbf{x}_{t}-\sqrt{1 - \alpha_{t}}\hat{\epsilon}_{t}}{\sqrt{\alpha_{t}}} \text{,}
\label{one_step_denoising}
\end{equation}
where scalar $\alpha_{t}$ governs denoising intensity at each timestep $t$. From Eq. ~\ref{one_step_denoising}, the approximation of the clean image $\hat{\mathbf{x}}_{0}$ can be predicted by performing one-step diffusion sampling.

\textbf{Diffusion guidance.} The purpose of diffusion guidance is to produce a more reasonable denoising direction based on auxiliary information $\mathbf{y}'$ during inference time \cite{bansal2023universal}. Specifically, at each timestep $t$, diffusion guidance enables the denoised image $ \mathbf{x}_{t-1}$ follow the auxiliary information $\mathbf{y}'$ by modifying the predicted noise $\hat{\epsilon}_{t}$ using the gradient of the energy function $ g(\mathbf{x}_{t};t,\mathbf{y}')$:
\begin{equation}
\tilde{\epsilon}_{t}=\hat{\epsilon}_{t}+\sigma_{t}\nabla_{\mathbf{x}_{t}}g(\mathbf{x}_{t};t,\mathbf{y}') \text{,}
\end{equation}
where $\sigma_{t}$ is a weighting schedule. 

In practice, the function $ g$ could be the approximate energy from any differentiable model, such as probabilities from a classifier \cite{dhariwal2021diffusion}, similarity scores from a vision-language model \cite{radford2021learning}, position penalties from a detection  \cite{cheng2024yolo} or segmentation model \cite{wang2022yolomask}, and motion consistencies from a flow estimator \cite{teed2020raft}. Recent studies \cite{bansal2023universal,gengmotion} have shown that the predicted clean image $\hat{\mathbf{x}}_{0}$ (See Eq. \ref{one_step_denoising}) can be straightforwardly used as the input for the energy function $g$, demonstrating that we do not require to learn an additional timestep-dependent differentiable model. This paper also computes the energy based on a one step approximate of the clean image $\hat{\mathbf{x}}_{0}$ to produce the gradient $\nabla_{\mathbf{x}_{t}}g(\mathbf{\hat{x}}_{0}; \mathbf{y}')$.
\section{Methodology}
\label{sec:methodology}
Our framework is designed for modular conditional image synthesis, generating diverse visual scenarios by flexibly combining three fundamental condition units: text, layout, and drag. Accordingly, it contains three diffusion guidance modules specifically designed for each unit. To clarify each module, we organize the methodology section as follows: in Sec.~\ref{sec:guidance_1}, we design a Dense Concept Alignment (DCA) module to make synthesized images follow semantic cues embedded in texts; in Sec.~\ref{sec:guidance_2}, we present a Dense Geometry Alignment (DGA) to enforce adherence to spatial configurations within layouts; in Sec.~\ref{sec:guidance_3}, we propose a Dense Motion Alignment (DMA) to allow diffusion model to be aware of movement and deformation. 

\subsection{Dense Concept Alignment (DCA)}
\label{sec:guidance_1}
\begin{figure}[t]
    \centering
    \includegraphics[width=0.48\textwidth]{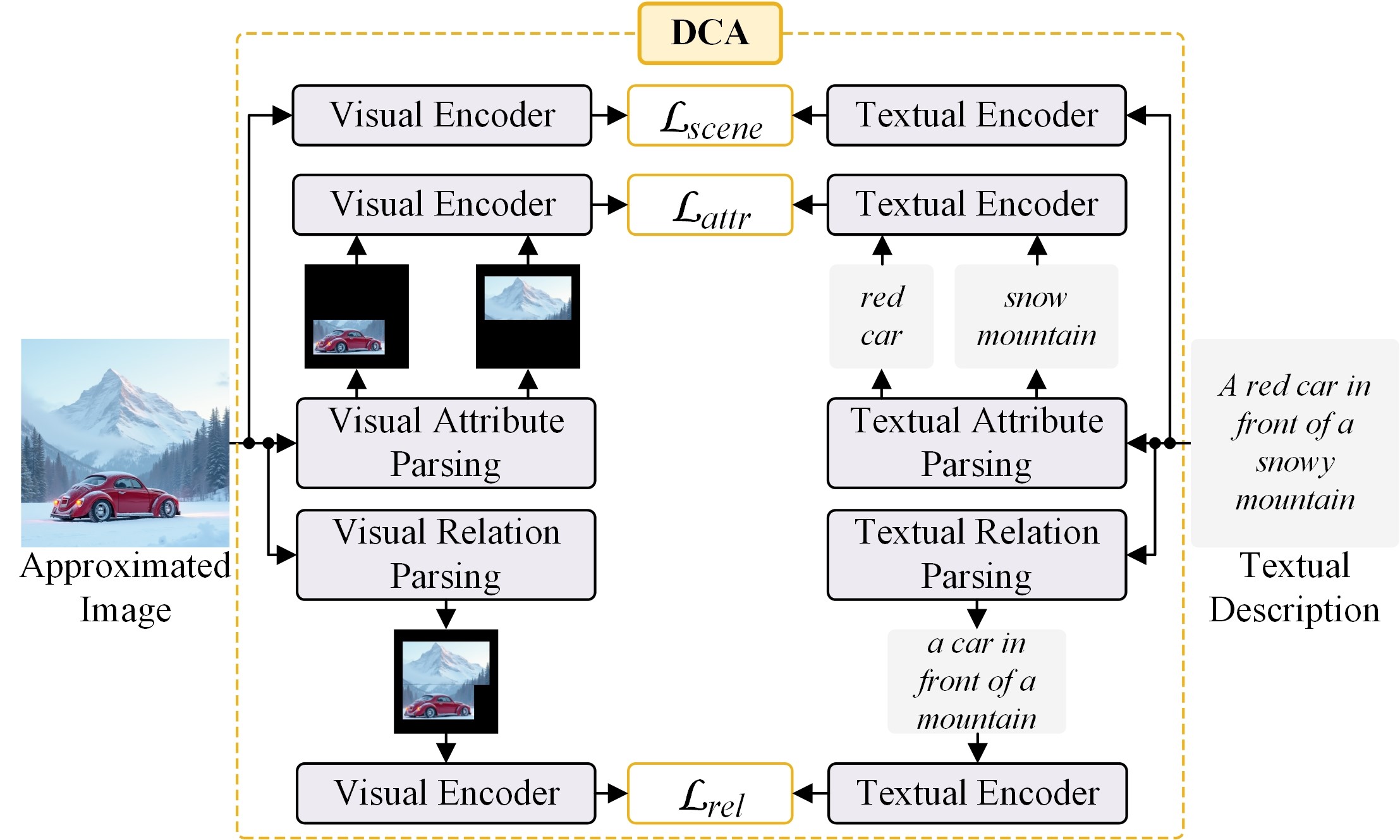} 
    \caption{The overall pipeline of our Dense Concept Alignment (DCA) module. Beyond scene-level vision-language consistency, this module can guide synthesized image in achieving dense alignment of attribute and relation concepts within textual conditions.}
    \label{FIG4_1}
\end{figure}

Given a textual description $\mathcal{T}$ as the condition, our goal is to guide the synthesized image to faithfully reflect its semantic cues. Current approaches \cite{wallace2023end} often regard the description $\mathcal{T}$ and predicted clean image $\hat{\mathbf{x}}_{0}$ as a whole for alignment. However, they fail to match compositional region and word information, resulting in semantic drift. Thus, we propose Dense Concept Align (DCA). As shown in Fig. \ref{FIG4_1}, we decouple $\mathcal{T}$ and $\hat{\mathbf{x}}_{0}$ into intra-object attributes and inter-object relations, and then align these concepts to establish a fine-grained vision-language correspondence.

We begin by deriving textual concepts from the description $\mathcal{T}$. Given an example like “A red car in front of a snowy mountain,” we employ scene graph parsing \cite{wu2019unified} to decompose it into textual attribute concepts (\textit{e.g.}, red car, snowy mountain) and relation concepts (\textit{e.g.}, car in front of mountain). Each concept is individually embedded using the CLIP text encoder \cite{radford2021learning}. An attribute concept is defined as an adjective-noun pair, where the adjective describes the property (\textit{e.g}., red, snowy) and the noun specifies the category (\textit{e.g}., car, mountain). We construct its embedding $\mathbf{w}^{\rm a,c}$ by concatenating the individual embeddings of the property and category: $\mathbf{w}^{\rm a,c} = \mathbf{w}^{\rm a} \oplus \mathbf{w}^{\rm c}$. In contrast, a relation concept is treated as a complete phrase and directly encoded to obtain its contextual representation $\mathbf{w}^{\rm r}$. This process yields a collection of textual concepts, denoted as $\mathcal{W}$:
\begin{equation}
\mathcal{W}=\{\{\mathbf{w}^{\rm a,c}_{i}\}_{i=1}^{I}, \{\mathbf{w}^{\rm r}_{k}\}_{k=1}^{K}\} \text{,}
\end{equation}
where $I$ and $K$ denote the number of parsed objects and relations, respectively.

Simultaneously, we utilize visual region parsing to extract visual concepts from image $\hat{\mathbf{x}}_{0}$. Concretely, we decouple the corresponding visual region of textual concepts using an open-set detection model \cite{liu2024grounding}. The visual attribute concept is represented by the detected bounding box enclosing an individual object, whereas the relation one is defined by the union of two object boxes. Using these bounding boxes, we crop $\hat{\mathbf{x}}_{0}$ to generate a set of isolated visual regions, which are then independently embedded by CLIP vision encoder\cite{radford2021learning}, forming the visual concept collection $\mathcal{V}$:
\begin{equation}
\mathcal{V}=\{\{\mathbf{v}^{\rm a,c}_{i}\}_{i=1}^{I}, \{\mathbf{v}^{\rm r}_{k}\}_{k=1}^{K}\} \text{,}
\end{equation}
where $\mathbf{v}^{\rm a,c}$ and $\mathbf{v}^{\rm r}$ represent the visual embeddings of attribute and relation concepts, respectively.

After obtaining embeddings of decoupled textual and visual concepts, beyond scene-level, we achieve finer-grained alignment by:
\begin{equation}
\mathcal{L}_{\rm DCA }=(1-\gamma) \mathcal{L}_{scene}+\gamma(\mathcal{L}_{attr} + \mathcal{L}_{rel}) \text{,}
\end{equation}
where
\begin{equation}
\mathcal{L}_{scene}={\rm cos}(\mathbf{w}^{\rm s},\mathbf{v}^{\rm s}) \text{,}
\end{equation}
\begin{equation}
\mathcal{L}_{attr}(\mathcal{W}, \mathcal{V})=\frac{1}{I}\sum_{i=1}^{I}{\rm cos}(\mathbf{w}^{\rm a,c}_{i},\mathbf{v}^{\rm a,c}_{i}) \text{,}
\end{equation}
\begin{equation}
\mathcal{L}_{rel}(\mathcal{W}, \mathcal{V})=\frac{1}{K}\sum_{k=1}^{K}{\rm cos}(\mathbf{w}^{\rm r}_{k},\mathbf{v}^{r}_{k}) \text{,}
\end{equation}
where ${\rm cos(\cdot, \cdot)}$ is the cosine similarity function. $\mathbf{w}^{\rm s}$ and $\mathbf{v}^{\rm s}$ are the global embeddings of description $\mathcal{T}$ and approximated image $\hat{\mathbf{x}}_{0}$, respectively. We set $\gamma=0.3$.

\subsection{Dense Geometry Alignment (DGA)}
\label{sec:guidance_2}
\begin{figure}[htbp]
    \centering
    \includegraphics[width=0.48\textwidth]{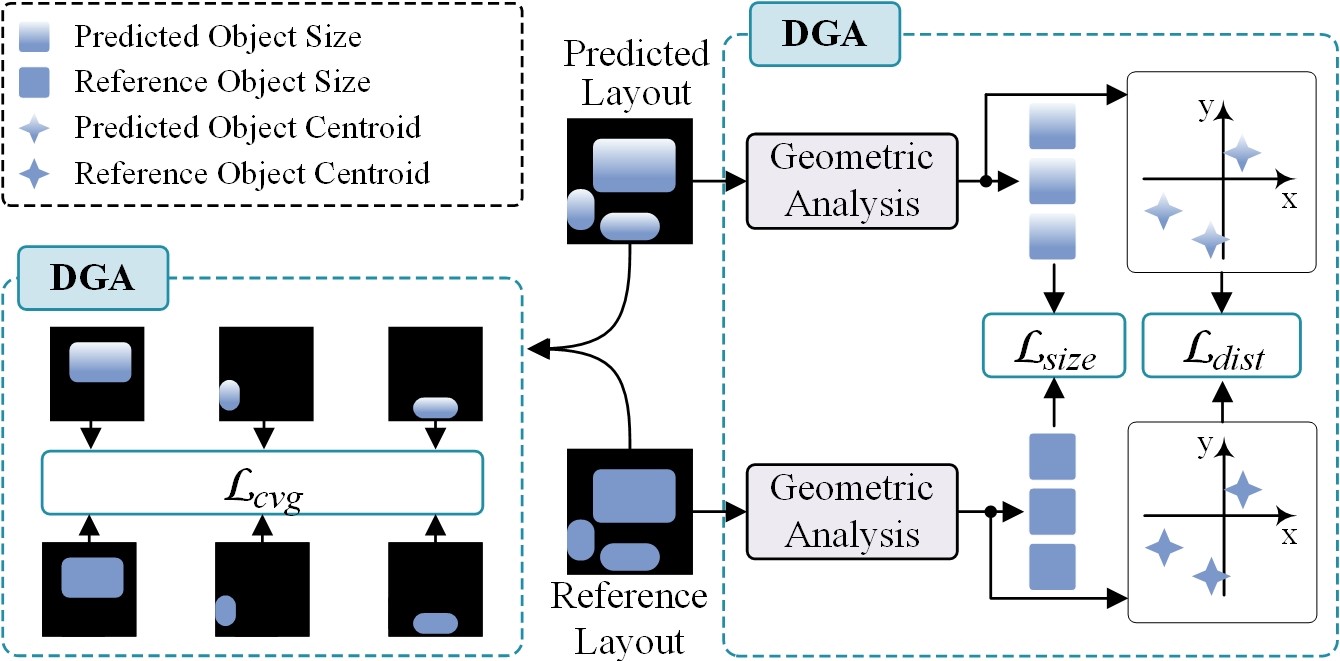} 
    \caption{The overall pipeline of our Dense Geometry Alignment (DGA) module. On the basis of the detection information from the approximated image, this module can densely align geometric features between the predicted and reference layouts, focusing on object-level location, size, and distance.}
    \label{FIG_4_2}
\end{figure}

To guide the synthesized image to resemble the spatial configuration described by the collection of segmentation masks $\mathcal{M}$ or 
bounding boxes $\mathcal{B} $, we propose Dense Geometry Alignment (DGA) module, as shown in Fig. \ref{FIG_4_2}. Current approaches simply ensure that
each object appears in its designated region by enhancing corresponding attention scores \cite{xie2023boxdiff,xiao2023r,kim2023dense,phung2024grounded}. Different from them, our module improves the generation plausibility from the perspectives of the object-wise locations, as well as relative size and distance between different objects. For simplicity, we use the segmentation mask collection $\mathcal{M}=\{\mathbf{m}\}_{n=1}^{N}$ as an example to illustrate this module.

Given the segmentation mask collection $\mathcal{\hat{M}}=\{\mathbf{\hat{m}}\}_{n=1}^{N}$ predicted from the approximated clean image $\hat{\mathbf{x}}_{0}$ by leveraging an image segmentation model \cite{wang2023yolov7}, considering an object with the $n$-th class, we strive to avoid it overlapping with the other objects and ensure the coverage of the target segmentation mask $\mathbf{m}_{n}$. Specifically, we enhance the IoU between the two segmentation masks $\mathbf{\hat{m}}_{n}$ and $\mathbf{m}_{n}$, and minimize the interaction between $\mathbf{\hat{m}}_{n}$ and other irrelevant masks $\mathcal{M}-\{\mathbf{m}_{n}\}$. Hence, the coverage loss function is formulated by: 
\begin{align}
\mathcal{L}_{cvg}(\mathcal{\hat{M}},\mathcal{M})=\sum_{n=1}^{N}(1 - \frac{\mathbf{\hat{m}}_{n} \cap \mathbf{m}_{n}}{{\mathbf{\hat{m}}_{n}} \cup \mathbf{m}_{n}}) \nonumber \\ +\sum_{n_{1}=1}^{N}\sum_{n_{2}=1}^{N}(\mathbf{\hat{m}}_{n_{1}} \cap \mathbf{m}_{n_{2}}),  
\end{align}
where for clarity and brevity, when two segmentation masks are concurrently referenced in an equation, we use the subscript $ n_{1}$ and $n_{2}$ to describe the class of different boxes for distinguishment.

Since maintaining realistic proportions between different objects is crucial for generating visually plausible images, we also restrict the pairwise size ratios of segmentation masks by the proposed size loss. On the basis of the areas computed for the predicted and target segmentation masks
, the size loss is formulated by: 
\begin{equation}
\mathcal{L}_{size}=\sum_{n_{1}=1}^{N}\sum_{n_{2}=1}^{N}(\frac{\mathbf{\hat{s}}_{n_{1}}}{\mathbf{\hat{s}}_{n_{2}}}-\frac{\mathbf{ {s}}_{n_{1}}}{\mathbf{ {s}}_{n_{2}}})^{2},
\end{equation} 
where $\mathbf{\hat{s}}_{n_{1}}$ ($\mathbf{\hat{s}}_{n_{2}}$) and $\mathbf{s}_{n_{1}}$ ($\mathbf{s}_{n_{2}}$) are areas of the segmentation masks in the collections $\mathcal{\hat{M}}$ and $\mathcal{M}$, respectively.

In addition, we enforce the alignment of pairwise distances between multiple objects among the approximated image $\hat{\mathbf{x}}_{0}$ and the given guidance information $\mathcal{M}$. In particular, after obtaining centroids of each predicted and target object
according to the segmentation mask collections $\mathcal{\hat{M}}$ and $\mathcal{M}$ respectively, we compute the pairwise Euclidean distances within each collection of centroids. By minimizing discrepancy in distances between predicted and target centroid collections, we formulate the distance loss function as: 
\begin{equation}
\mathcal{L}_{dist}=\sum_{n_{1}=1}^{N}\sum_{n_{1}=2}^{N}(\| \mathbf{\hat{c}}_{n_{1}} - \mathbf{\hat{c}}_{n_{2}} \|_{2} - \|\mathbf{c}_{n_{1}} - \mathbf{c}_{n_{2}} \|_{2}),
\end{equation}
where $\mathbf{\hat{c}}_{n_{1}}$ ($\mathbf{\hat{c}}_{n_{2}}$) and $\mathbf{c}_{n_{1}}$ ($\mathbf{c}_{n_{2}}$) are centroids of the segmentation masks in the collections $\mathcal{\hat{M}}$ and $\mathcal{M}$, respectively.

In summary, our final guidance function for DGA can be defined as follows:
\begin{equation}
\mathcal{L}_{\rm DGA}=(1-\lambda) \mathcal{L}_{cvg}+\lambda(\mathcal{L}_{size}+\mathcal{L}_{dist}),
\end{equation}
where we set $\lambda=0.25$.

\subsection{Dense Motion Alignment (DMA)}
\label{sec:guidance_3}
\begin{figure}[t]
    \centering
    \includegraphics[width=0.48\textwidth]{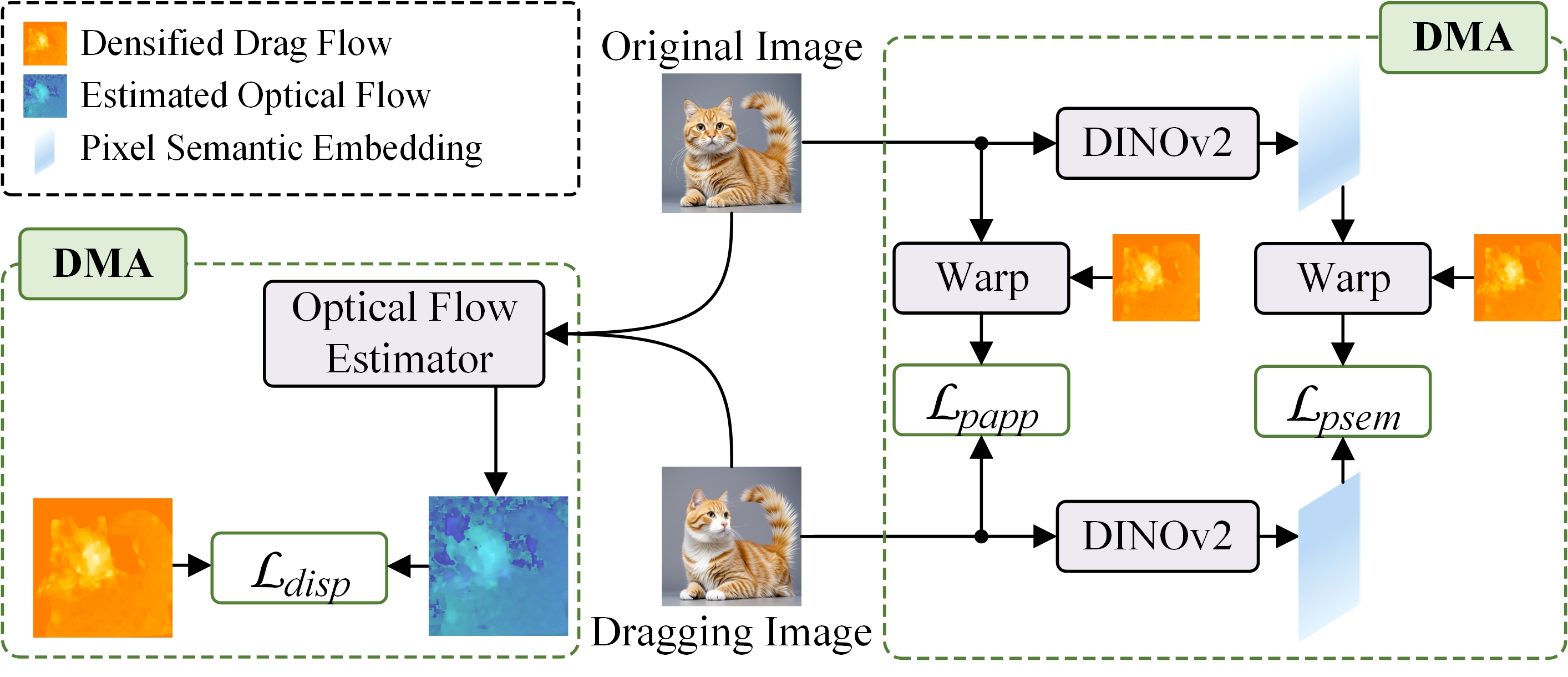} 
    \caption{The overall pipeline of our Dense Motion Alignment (DMA) module. Using our densified reference drag flow, this module aims to guide synthesized image in achieving dense alignment of motion information across drag conditions, encompassing pixel displacement, appearance, and semantic aspects.}
    \label{FIG4_3}
\end{figure}

Given a collection of $J$ drag signals $\mathcal{P}=\{(\mathbf{p}^{\rm o}_{j},\mathbf{p}^{\rm d}_{j})\}_{j=1}^{J}$, we aim to move visual contents of each original point $\mathbf{p}^{\rm o}_{j}$ to its destination $\mathbf{p}^{\rm d}_{j}$. Recent studies, \textit{e.g.}, DragDiffusion \cite{shi2024dragdiffusion} and DragNoise \cite{liu2024drag}, only regulate feature similarity around $\mathbf{p}^{\rm o}_{j}$ and $\mathbf{p}^{\rm d}_{j}$ by implicit optimization in one selected diffusion step. However, on one hand, sparse drag signals do not provide sufficient object awareness, causing incomplete displacement in foreground. On the other hand, one-step feature optimization is unstable. To enable continuous dragging of designated pixels while preserving visual coherence during diffusion sampling, we propose Dense Motion Alignment (DMA), as shown in Fig.~\ref{FIG4_3}. Specifically, we predict reference dense drag flow, and enforce pixel correspondence between an image pair in terms of displacement, appearance, and semantics building on this flow. We define $\hat{\mathbf{x}}_{0}^{t+1}$ and $\hat{\mathbf{x}}_{0}^{t}$ as original and dragging images, respectively, for each timestep $t$.

We generate the reference dense drag flow for each individual drag signal using an object-specific displacement field. In particular, we employ SAM \cite{kirillov2023segment} to obtain a segmentation mask collection $\{\mathbf{m}^{\rm sam}_{j}\}_{j=1}^{J}$, where each mask represents an instance area of $\mathbf{p}^{\rm o}_{j}$. Because image pixels closer to $\mathbf{p}^{\rm o}_{j}$ should have larger displacements \cite{wu2025draganything}, we use radial basis functions to produce a collection of 2D Gaussian maps $\{\mathbf{g}_{j}\}_{j=1}^{J}$. Each map depicts gradually diminishing drag influence radiating outward from $\mathbf{p}^{\rm o}_{j}$. Based on $\{\mathbf{m}^{\rm sam}_{j}\}_{j=1}^{J}$ and $\{\mathbf{g}_{j}\}_{j=1}^{J}$, our reference flow can be densified by:
\begin{equation}
\mathbf{u}_{\hat{\mathbf{x}}_{0}^{t+1} \rightarrow \hat{\mathbf{x}}_{0}^{t}, j}= (\mathbf{p}^{\rm d}_{j} - \mathbf{p}^{\rm o}_{j}) * \mathbf{m}^{\rm sam}_{j} * \mathbf{g}_{j}, 
\end{equation}
where $\mathbf{u}_{\hat{\mathbf{x}}_{0}^{t+1} \rightarrow \hat{\mathbf{x}}_{0}^{t}, j}$ is the densified drag flow for $j$-th signal.

To rectify discrepancies in pixel displacement direction and magnitude, we enable the estimated optical flow between $\mathbf{\hat{x}}_{0}^{t+1}$ and $\mathbf{\hat{x}}_{0}^{t}$ to align with its reference flow  by minimizing displacement loss:
\begin{equation}
\mathcal{L}_{disp}=\sum_{j=1}^{J}||\mathbf{\hat{u}}_{\mathbf{\hat{x}}_{0}^{t+1}\rightarrow{\mathbf{\hat{x}}}_{0}^{t},j} - \mathbf{u}_{\mathbf{\hat{x}}_{0}^{t+1}\rightarrow{\mathbf{\hat{x}}}_{0}^{t},j}||_{1},
\end{equation}
where $\mathbf{\hat{u}}_{\mathbf{\hat{x}}_{0}^{t+1}\rightarrow{\mathbf{\hat{x}}}_{0}^{t},j}$ is an optical flow estimated by a RAFT model \cite{teed2020raft}. In order to ensure visual coherence, we enforce pixel-level appearance consistency by directly minimizing the differences between the dragging and warped original image, while semantic consistency is preserved by DINO v2 embedding \cite{oquab2023dinov2}:
\begin{equation}
\mathcal{L}_{papp}=\sum_{j=1}^{J}\sqrt{({\rm warp}(\hat{\mathbf{x}}_{0}^{t+1}, \mathbf{u}_{\hat{\mathbf{x}}_{0}^{t+1}\rightarrow\hat{\mathbf{x}}_{0}^{t},j})-\hat{\mathbf{x}}_{0}^{t})^{2}},
\end{equation}
\begin{equation}
\mathcal{L}_{psem}=\sum_{j=1}^{J}\sqrt{({\rm warp}(\hat{\mathbf{f}}_{0}^{t+1}, \mathbf{u}_{\hat{\mathbf{x}}_{0}^{t+1}\rightarrow\hat{\mathbf{x}}_{0}^{t},j})-\hat{\mathbf{f}}_{0}^{t})^{2}},
\end{equation}
where $\hat{\mathbf{f}}_{0}^{t+1}$ and $\hat{\mathbf{f}}_{0}^{t}$ are semantic embedding calculated by DINO v2 model. For a drag signal, our loss of DMA module is defined as:
\begin{equation}
\mathcal{L}_{\rm DMA}=(1-\eta) \mathcal{L}_{disp} + \eta(\mathcal{L}_{papp}+\mathcal{L}_{psem}),
\end{equation}
where we set $\eta=0.02$. Like DragDiffusion \cite{shi2024dragdiffusion} and DragNoise \cite{liu2024drag}, we also update the positions of the original points after each denosing step.
\section{Experiments}
\subsection{Experimental Setups}
This section sequentially describe used datasets and evaluation metrics in our experiments, and implementation details of our approaches. 

\textbf{Datasets.} The datasets used in our study are: (1) COCO \cite{lin2014microsoft}. We curate the 400 longest image descriptions from COCO validation set, each ranging from 15 to 50 words. This selection is made to better demonstrate our strength in understanding long-form description; (2) DenseDiffusion \cite{kim2023dense}. This benchmark contains 250 samples (2$\sim$3 unique objects of each sample), where each sample has a global image description, a segmentation mask, and local descriptions and class labels for each segment; (3) DrawBench \cite{saharia2022photorealistic}. We use the `position' subset from it. This subset is composed of 20 samples (2 unique objects of each sample), where each sample has a description and some bounding boxes of each object (box coordinates and class labels generated by GPT); and (4) DragBench \cite{shi2024dragdiffusion}. This benchmark comprises 205 samples, each of which has a reference image, one or more drag signals, a description, and a mask indicating regions to be modified.

\textbf{Evaluation metrics.} For \textit{text} condition, we use  Fréchet Inception Distance (FID) \cite{heusel2017gans} and  CLIP score \cite{radford2021learning}. FID quantifies the similarity between the synthesized and real images, while the CLIP score measures the alignment between the synthesized image and the reference description. For \textit{text} $+$ \textit{layout} condition, we use IoU and SOA-I \cite{hinz2020semantic} (layout prediction is based on YOLOv7-Seg \cite{wang2023yolov7}). IoU quantifies the overlap between the predicted layout from the synthesized image and the reference layout. SOA-I measures whether described objects appear in the synthesized image. For \textit{text} $+$ \textit{drag} condition, we use Image Fidelity (IF) \cite{kawar2023imagic} and Mean Distance (MD) \cite{pan2023drag}. IF (1-LPIPS score \cite{zhang2018unreasonable}) measures the image quality based on the global similarity between the reference and dragged images. MD assesses how accurately  the post-dragging original point align with its corresponding destination point by measuring  Euclidean distance between their positions.

\textbf{Implementation details.} We use $T$=50 steps of the DDIM scheduler \cite{2021Denoising} and $T$=28 steps of the Flow Match Euler Discrete scheduler \cite{gat2024discrete}. The weighting schedules for our proposed DCA, DGA, and DMA modules are set to 60, 25, and 90, respectively. Our guidance is only applied during the first 70\% timesteps. 

\subsection{Image Synthesis Results with Text}
\begin{figure}[t]
    \centering
    \includegraphics[width=1.0\columnwidth]{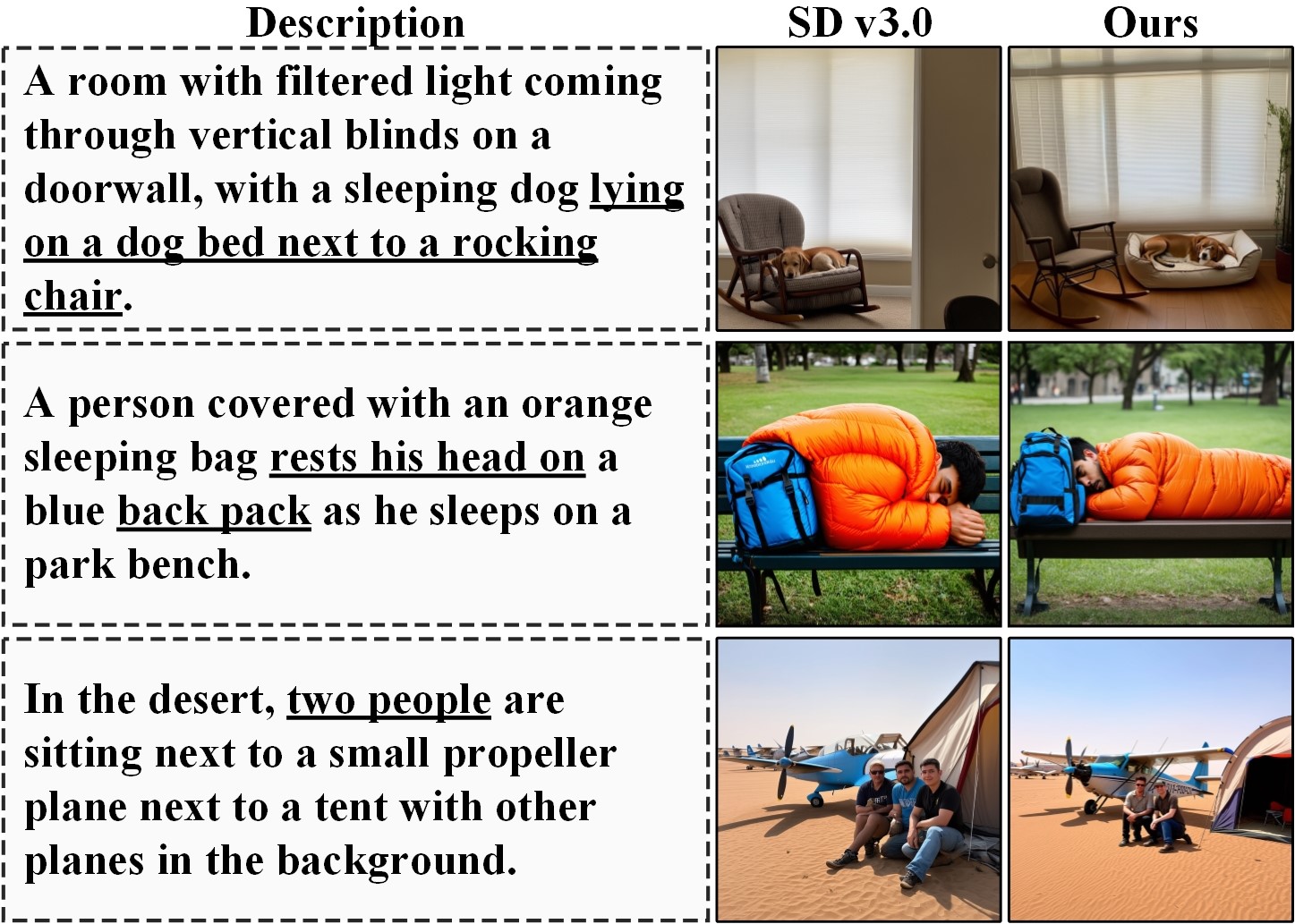} 
    \caption{Visualization comparison between our approach (Stable Diffusion v3.0 \cite{esser2024scaling} + DCA) and its baseline under varying descriptions.  Our approach demonstrates superior performance in aligning semantic concepts within descriptions.}
    \label{FIG5_1}
\end{figure}

We use Stable Diffusion v1.5 \cite{rombach2022high} and v3.0 \cite{esser2024scaling} as baseline models to evaluate the effectiveness of our DCA module in aligning concepts within descriptions. Tab.~\ref{TABLE5_1} summarizes FID \cite{heusel2017gans} and CLIP scores \cite{radford2021learning} on long ($>$15 words) descriptions from COCO dataset \cite{lin2014microsoft}, our DCA module achieves 12.3\% and 2.97\% improvements in FID and CLIP scores, respectively, building upon Stable Diffusion v3.0 (the most recent baseline). This gain arises from our module's ability to achieve vision-language alignment in a coarse-to-fine manner, effectively leveraging semantic cues embedded in complex sentences. Besides, our module can be applicable to Stable Diffusion v1.5, showing its adaptability across different baseline versions. As shown in Fig.~\ref{FIG5_1}, our approach can properly place the dog on a dog bed next to a rocking chair (top) and position the person’s head on a blue backpack (middle), indicating its effectiveness in aligning the object interaction concept. Our approach can also generate two people (bottom), showcasing its advantage in understanding the object's numeric information.
\begin{table}[t]
\centering
\caption{Quantitative comparison under long descriptions from COCO dataset \cite{lin2014microsoft}. The evaluation metrics used are FID \cite{heusel2017gans} and CLIPScore \cite{radford2021learning}. We reimplement all baseline approaches. 
The best results are \textbf{highlighted}.}
\label{TABLE5_1}
\resizebox{\columnwidth}{!}{ 
\begin{tabular}{cccc}
\toprule[1.5pt]
Methods & Venue & FID $ \downarrow $ & CLIPScore $ \uparrow $  \\
\midrule
ParaDiffusion \cite{wu2023paragraph} & arXiv2023 & 12.96 & 0.2731 \\
PIXART-$\alpha$ \cite{chenpixart} & ICLR2024 & 13.97 & 0.2724 \\
SDXL \cite{podellsdxl} & ICLR2024 & 9.79 & 0.2820 \\
ELLA \cite{hu2024ella} & arXiv2024 & 13.64 & 0.2811 \\
PIXART-$\Sigma$ \cite{chen2024pixart} & ECCV2024 & 10.73 & 0.2793 \\
\midrule
Stable Diffusion v1.5 \cite{rombach2022high} & CVPR2022 & 11.49 & 0.2750  \\
+ DCA & -  & 10.29 & 0.2882 \\
\midrule
Stable Diffusion v3.0 \cite{esser2024scaling} & ICML2024 & 10.51 & 0.2861  \\
+ DCA & -  & \textbf{9.22} & \textbf{0.2946} \\
\bottomrule[1.5pt]
\end{tabular}
}
\end{table}

\subsection{Image Synthesis Results with Layout and Text}
\begin{figure}[t]
    \centering
    \includegraphics[width=\columnwidth]{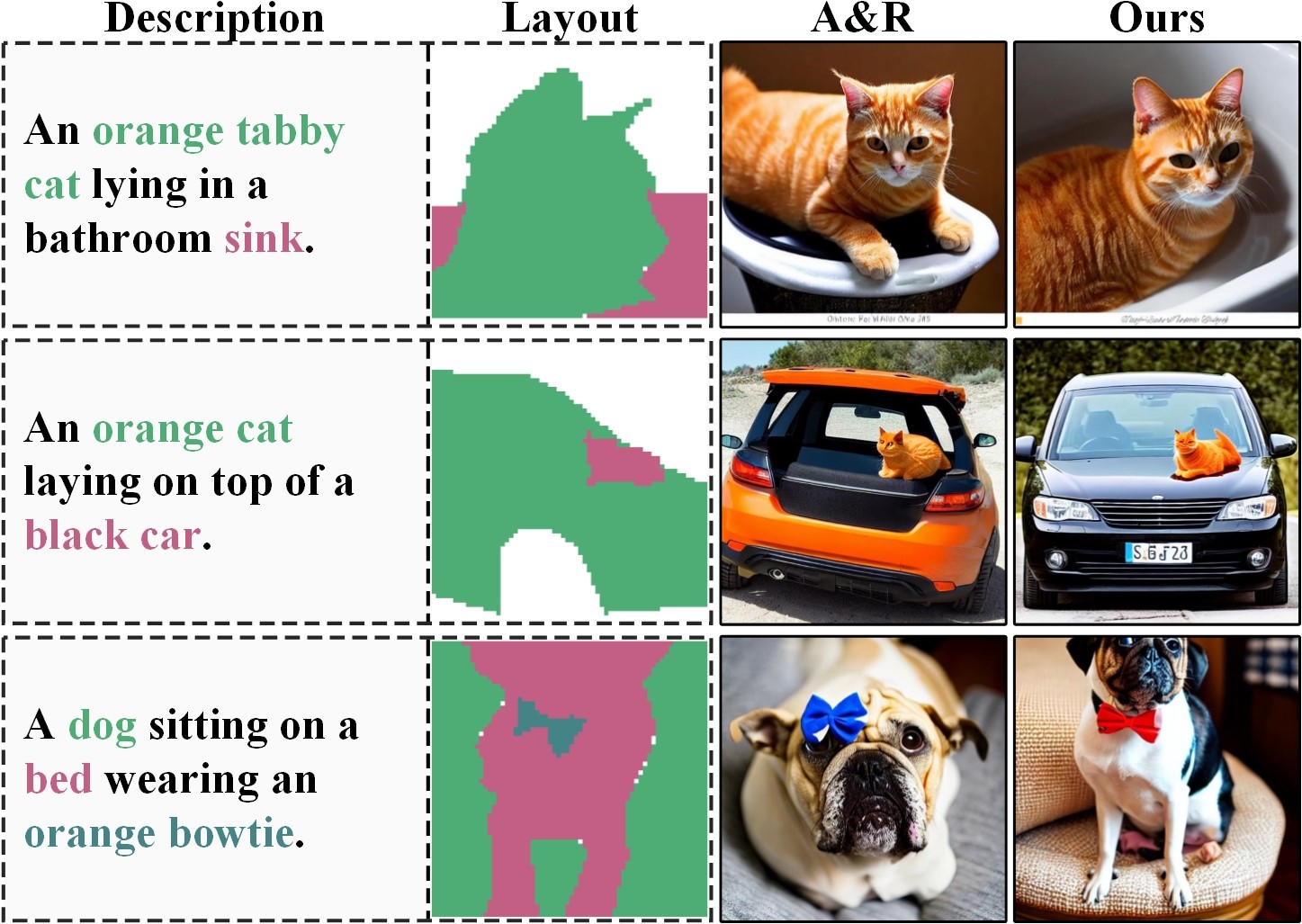} 
    \caption{Visualization comparison between our approach (A\&R \cite{phung2024grounded} + DGA + DCA) and its baseline, under varying descriptions and segmentation masks. Our approach can adhere to both geometric concepts and positional configuration.}
    \label{FIG5_2}
\end{figure}

\begin{figure}[t]
    \centering
    \includegraphics[width=\columnwidth]{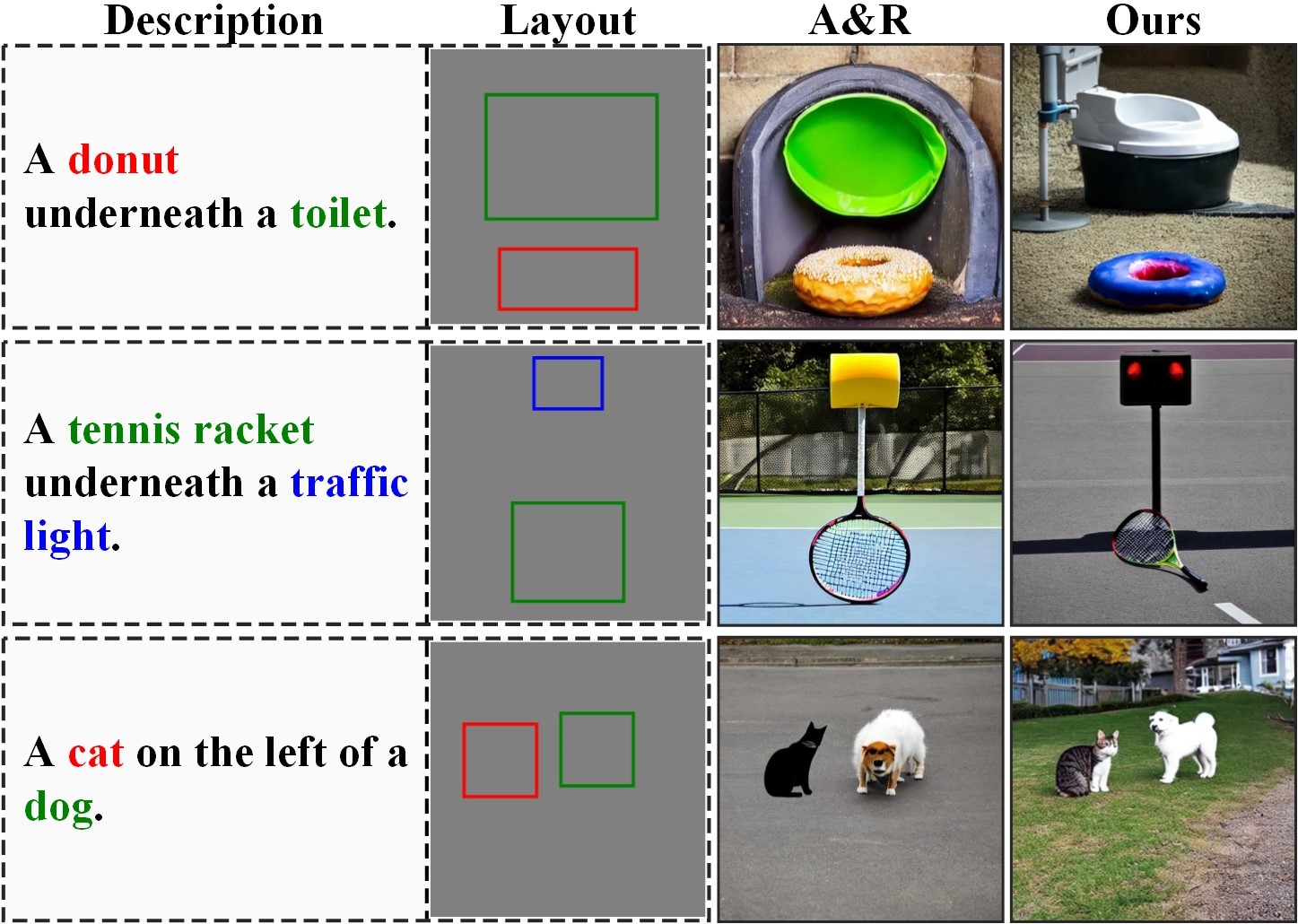} 
    \caption{Visualization comparison between our approach (A\&R \cite{phung2024grounded} + DGA + DCA) and its baseline, under varying descriptions and bounding boxes. Our approach can satisfy both conditions.}
    \label{FIG5_3}
\end{figure}

We use Stable Diffusion v1.5 \cite{rombach2022high}, DenseDiffusion \cite{kim2023dense} and A\&R \cite{phung2024grounded} as baseline models to evaluate the effectiveness of our DGA module in aligning spatial configurations represented by segmentation masks and bounding boxes; and our DCA module in aligning geometric concepts within descriptions. Tab.~\ref{TABLE_2} summarizes IoU and SOA-I scores on mask-to-image. Our DGA module brings a 27.1\% improvement in IoU and a 7.55\% increase in SOA-I over A\&R baseline, owing to the establishment of object-wise geometric consistency. Besides, incorporating our DCA module on top of DGA leads to an additional improvement, achieving an IoU of 50.61 and an SOA-I of 86.33. This improvement is due to the ability of DCA module to understand descriptions of positions and shapes. We can also observe consistent performance gains across both Stable Diffusion v1.5 and DenseDiffusion architectures. As shown in Fig.~\ref{FIG5_2}, our approach can accurately place a cat inside a bathroom sink (top), position a cat on top of a black car (middle), generate a bowtie around a dog’s neck (bottom), demonstrating the advantage of our framework in awareness of spatial relationships. The same conclusion can be observed in Fig.~\ref{FIG5_3}, which presents box-to-image visualizations of validation data from the DrawBench benchmark \cite{saharia2022photorealistic}.
\begin{table}[t]
\centering
\caption{Quantitative comparison under descriptions and segmentation masks provided by DenseDiffusion \cite{kim2023dense}. The evaluation metrics are IoU and SOA-I \cite{hinz2020semantic}. We generate four random images of each condition combination for performance evaluation. The best results are \textbf{highlighted}.}
\label{TABLE_2}
\resizebox{\columnwidth}{!}{ 
\begin{tabular}{cccc}
\toprule[1.5pt]
Methods  & Venue & IoU $ \uparrow $ & SOA-I  $ \uparrow $ \\
\midrule
Real Images  & - & 59.13 & 93.07 \\
\midrule
Composable Diffusion \cite{liu2022compositional} & ECCV2022 & - & 55.88$\pm$0.78 \\ 
Structure Diffusion \cite{feng2023training} & ICLR2023 & - & 70.97$\pm$1.08 \\ 
SD-Pww \cite{balaji2022ediff} & arXiv2022 & 23.76$\pm$0.50 & 73.92$\pm$1.84 \\ 
\midrule
Stable Diffusion \cite{rombach2022high} & CVPR2022  & - & 73.08$\pm$1.54 \\ 
+ DGA & - & 25.46$\pm$1.02 & 74.45$\pm$1.46 \\
+ DGA + DCA & - &  27.28$\pm$0.97 & 77.02$\pm$1.67 \\
\midrule
Dense Diffusion \cite{kim2023dense} & ICCV2023 & 34.99$\pm$1.13 & 77.61$\pm$1.75 \\ 
+ DGA & - & 36.72$\pm$1.24 & 78.21$\pm$1.62 \\
+ DGA + DCA & - & 38.06$\pm$1.26 & 78.42$\pm$1.17 \\  
\midrule
A\&R \cite{phung2024grounded} & CVPR2024 & 38.97$\pm$0.56 & 78.80$\pm$1.27 \\
+ DGA & - & 49.52$\pm$0.70 & 84.75$\pm$2.46 \\
+ DGA + DCA & - & \textbf{50.61$\pm$0.47} & \textbf{86.33$\pm$0.88} \\ 
\bottomrule[1.5pt]
\end{tabular}}
\end{table}
\begin{figure}[!htbp]
    \centering
    \includegraphics[width=\columnwidth]{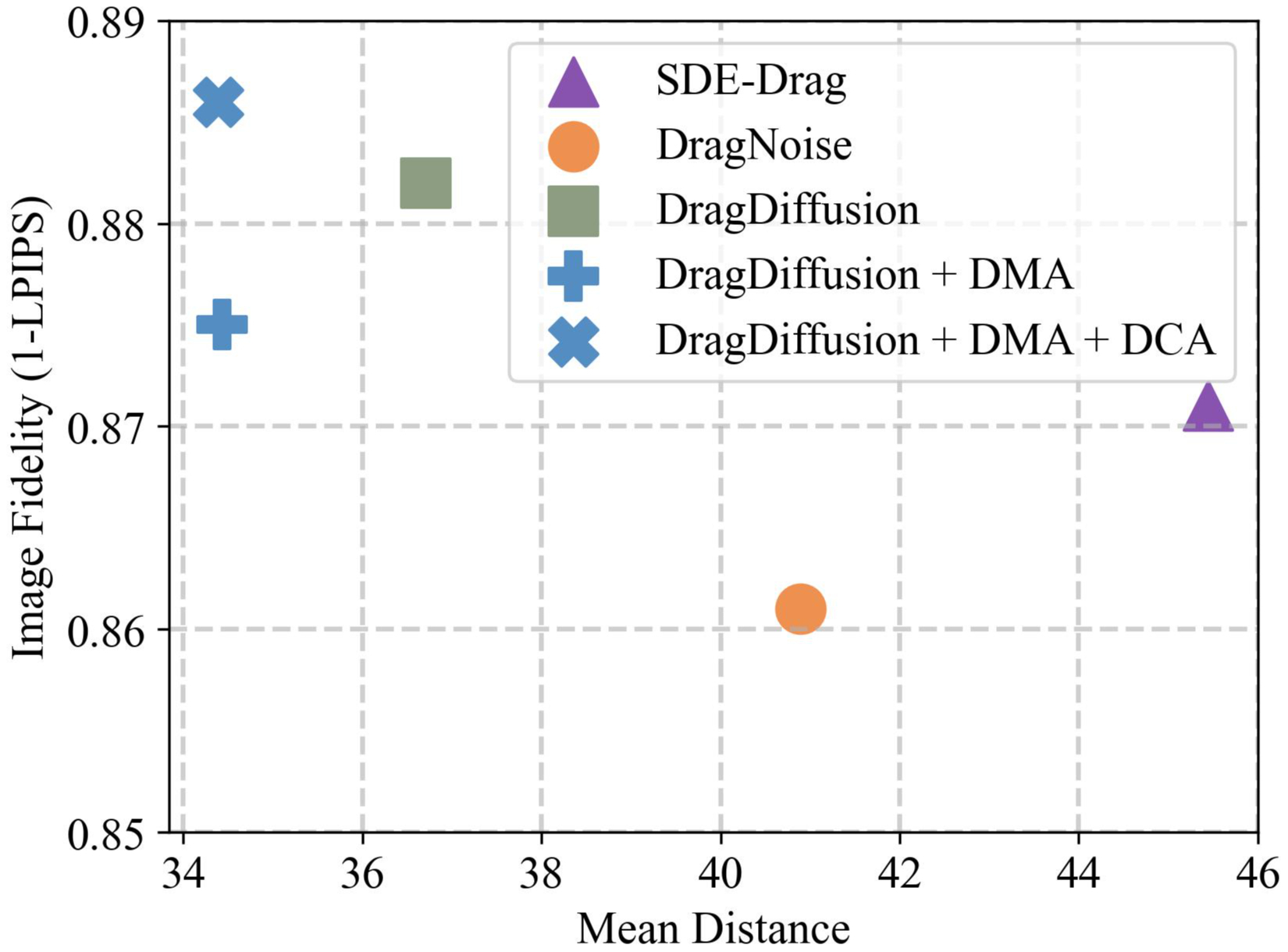} 
    \caption{Quantitative comparison under descriptions and drag signals from DragBench benchmark \cite{shi2024dragdiffusion}. The evaluation metrics used are Mean Distance ($\downarrow$) and Image Fidelity (1-LPIPS) ($\uparrow$). Approaches with the best results appear in the upper-left corner.}
    \label{FIG5_3.9}
\end{figure}

\subsection{Image Synthesis Results with Drag and Text}
\begin{figure}[!t]
    \centering
    \includegraphics[width=\columnwidth]{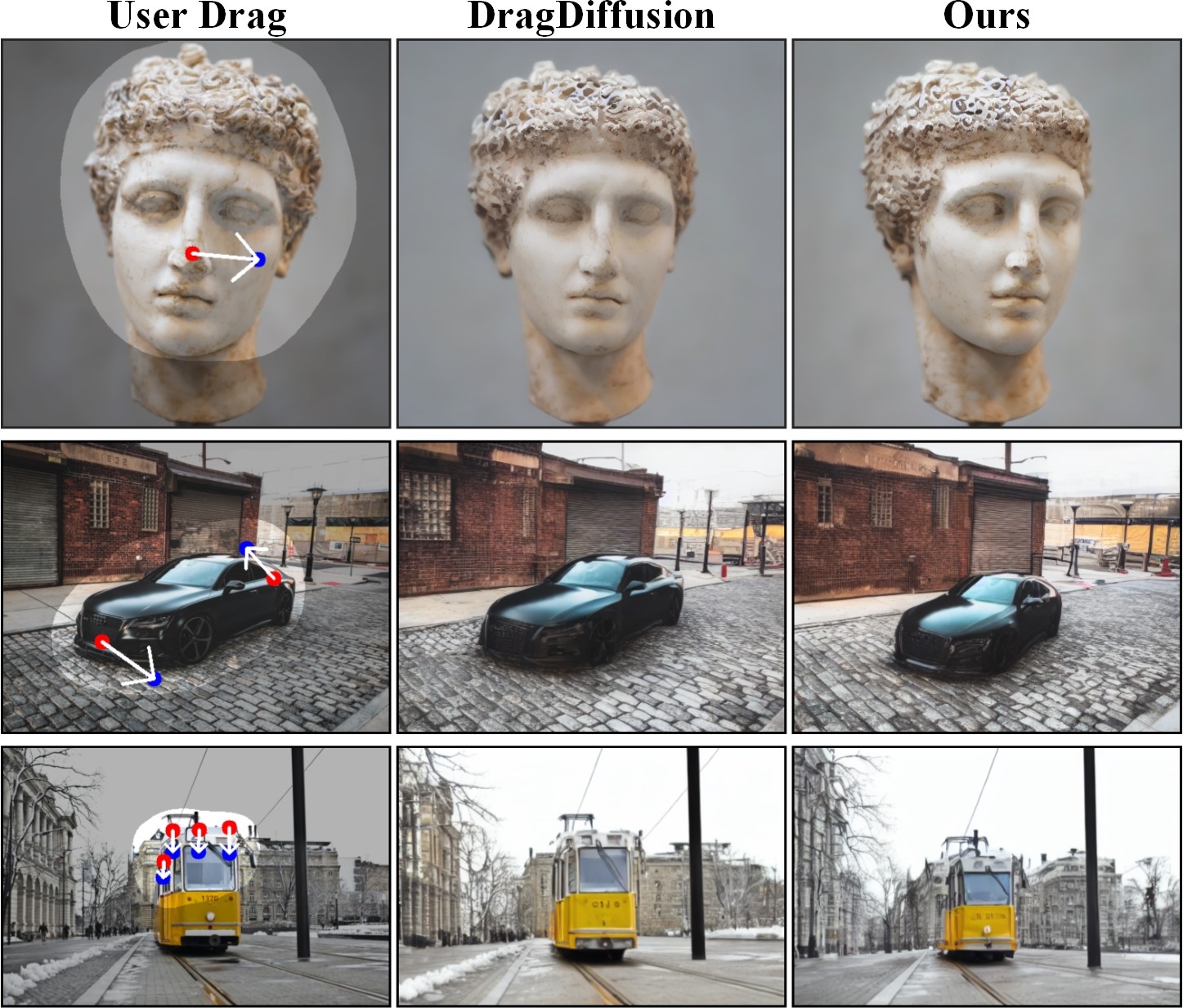} 
    \caption{Visualization comparison between our approach (DragDiffusion \cite{shi2024dragdiffusion} + DMA) and its baseline under varying drag signals. Our approach enables effective synthesis of large-scale movement and deformation.}
    \vspace{-3mm}
    \label{FIG5_4}
\end{figure}

We use DragDiffusion \cite{shi2024dragdiffusion} as a baseline model to evaluate the effectiveness of our DMA module in motion alignment; and our DCA module for semantic preservation. Fig.~\ref{FIG5_3.9} presents MD and IF scores on the DragBench benchmark \cite{shi2024dragdiffusion}. The combination of our DMA and DCA modules achieves better MD and IF scores, indicating that our framework ensures each pixel follows its desired trajectory with fewer visual artifacts. However, applying our DMA module alone results in a slight drop in IF, compared to DragDiffusion. This is because our motion field may cause over-smoothing in detail. Fortunately, our DCA module can mitigate such an issue to some degree, by maintaining the overall semantic structure unchanged during the diffusion sampling process. As shown in Fig.~\ref{FIG5_4}, our framework can effectively synthesize head rotation (top), rigid car displacement (middle), and large-scale deformation of the bus (bottom). This demonstrates that our algorithm enables the desired movement even when handling complex drag signals (multiple drag points).
\section{Conclusion}
\begin{figure}[htb]  
    \centering
    \includegraphics[width=1.0\columnwidth]{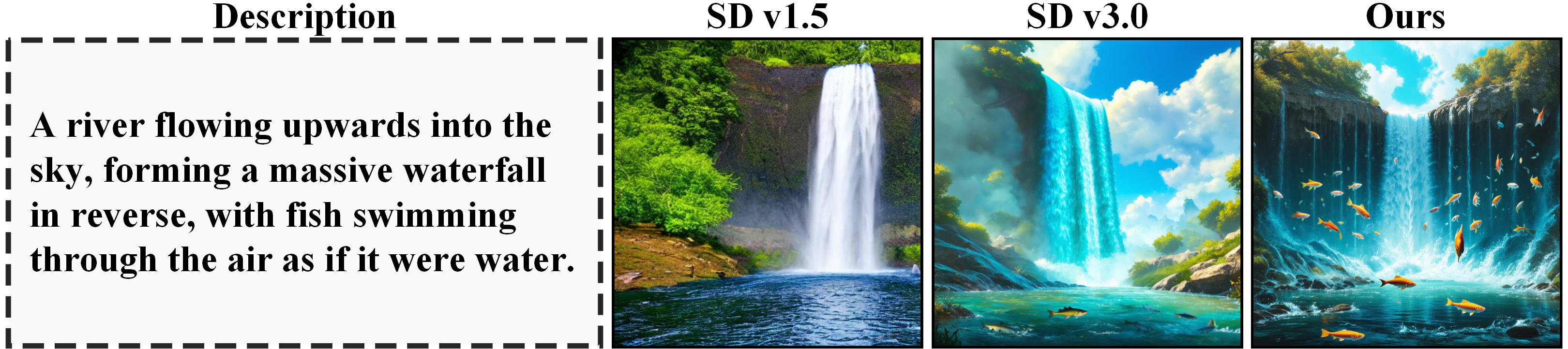}
    \caption{Failure cases. As foundation model lacks world knowledge, our approach (Stable Diffusion v3.0 \cite{esser2024scaling} + DCA) fails to generate counterfactual scenes.}
    \label{failure}
\end{figure}

We propose a plug-and-play modular design for guiding image synthesis by aligning diverse fundamental condition units (text, layout, drag) in combination. Based on this idea, we devise three dense alignment approaches to independently achieve flexible control over concept, geometry and motion. Comprehensive experiments demonstrate the effectiveness of our framework across diverse conditioning scenarios. However, our framework relies on the capability of its foundation model, \textit{e.g.}, SD model. As shown in Fig.~\ref{failure}, counterfactual scenes is hard to generate, \textit{e.g.}, a river flowing upwards, if SD model is unable to synthesis this scene itself. Future work could explore world knowledge-aware generative paradigms to address this challenge.

\noindent \textbf{Acknowledgement:} This work was supported by the National Natural Science Foundation of China (No. U23B2013, 62276176). This work was also partly supported by the SICHUAN Provincial Natural Science Foundation (No. 2024NSFJQ0023).

\clearpage
{
    \small
    \bibliographystyle{ieeenat_fullname}
    \bibliography{main}
}

\end{document}